\documentclass[10pt,twocolumn,letterpaper]{article}

\usepackage{iccv}
\usepackage{times}
\usepackage{epsfig}
\usepackage{graphicx}
\usepackage{amsmath}
\usepackage{amssymb}
\usepackage{subfig}

\newcommand{\note}[1]{}%\textcolor{red}{note: #1}\PackageWarning{note:}{#1!}}
\newcommand{\margin}{Max-Margin Loss\xspace}

\usepackage[pagebackref=true,breaklinks=true,letterpaper=true,colorlinks,bookmarks=false]{hyperref}

\iccvfinalcopy % *** Uncomment this line for the final submission

\def\iccvPaperID{15} % *** Enter the ICCV Paper ID here

\ificcvfinal\fi
\begin{document}

%%%%%%%%% TITLE
\title{Learning Deep Convolutional Embeddings for Face Representation Using Joint Sample- and Set-based Supervision}

\author{Baris Gecer,  Vassileios Balntas, and  Tae-Kyun Kim\\
Department of Electrical and Electronic Engineering,\\
Imperial College London\\
{\tt\small $\{$b.gecer,v.balntas15,tk.kim$\}$@imperial.ac.uk}
%\and
%{\tt\small tk.kim@imperial.ac.uk}
}

\maketitle
%\thispagestyle{empty}

%%%%%%%%% ABSTRACT
\begin{abstract}
   In this work, we investigate several methods and strategies to learn deep embeddings for face recognition, using joint sample- and set-based optimization. We explain our framework that expands traditional learning with set-based supervision together with the strategies used to maintain set characteristics. We, then, briefly review the related set-based loss functions, and subsequently propose a novel \margin which maximizes maximum possible inter-class margin with assistance of Support Vector Machines (SVMs). It implicitly pushes all the samples towards correct side of the margin with a vector perpendicular to the hyperplane and a strength exponentially growing towards to negative side of the hyperplane. We show that the introduced loss outperform the previous sample-based and set-based ones in terms verification of faces on two commonly used benchmarks. 
   
\end{abstract}

%-------------------------------------------------------------------------
\section{Introduction}
\label{sec:intro}
%Convolutional Neural Networks (CNN) are a visual form \note{visual form?} of hierarchical networks where discriminative, representative features are learned for end-to-end recognition from raw pixels of the original image to class labels. \note{this paragraph is too general I think its not needed - or if you want to keep it in, maybe add references and specific examples.}

Recently, deep convolutional neural networks (CNNs) have been an important tool that achieves state-of-the-art performances in many computer vision tasks \cite{lecun2015deep}. Its goal is to build a model to address a target problem with a sequence of convolutional layers that are developing from low-level features to more abstract representations. Deep networks can also learn robust representations that are suitable for other task \cite{donahue2014decaf, sharif2014cnn, qian2015fine, snoek2015scalable,gecer2016detection}. Deep Distance Metric Learning (DML) approaches explore ways to construct such representations that maintain better similarity/distance measurement for, \eg, verification, retrieval or clustering tasks. While supervision of traditional objective functions (\eg Softmax Loss) yield successful results, comparative loss functions (i.e. Triplet Loss) are shown to be more suitable for semi-supervised deep DML tasks\cite{parkhi2015deep}. 

Beside sample-based supervision which processes each sample individually, one can benefit from the captured information by considering a set of images as a unified entity. An image set is a collection of instances of the same object/person from varying viewpoints, illuminations, poses and exhibits different characteristics. A set contains richer information of the target than a single image and is potentially more useful for problems like object or scene classification, face recognition and action analysis. As the authors of \cite{wen2016discriminative} illustrated, set-based supervision can learn discriminative features rather than just separable features like sample-based approaches would learn.
 
This paper makes the following contributions:

\begin{itemize}
\item We propose a novel loss function called \margin that benefits from set-based information by drawing inter-set (inter-class) margins. It improves the separability of learned features by maximizing the maximum possible inter-class margin that is calculated by a support vector machine and address the shortcomings of the existing set-based methods.
  
\item We review existing set-based DML approaches and evaluate them and their combinations together with \margin and Softmax Loss.
  
\item We build a framework where such functions can operate properly jointly with sample-based ones and investigate the strategies to maintain set information during training in the framework.

\end{itemize}

The rest of the paper is organized as follows: In Section 2, we provide an overview of the related work about sample-,set-based deep metric learning for face recognition. Section 3 describes existing and new set-based loss functions and other strategies. In Section 4, we provide some more information on the set-up and techniques used in the experiments. We then present and discuss some experimental results. Finally, we draw conclusion and elaborate on future works in Section 5.

\begin{figure*}[t]
\begin{center}
\includegraphics[width=1\linewidth]{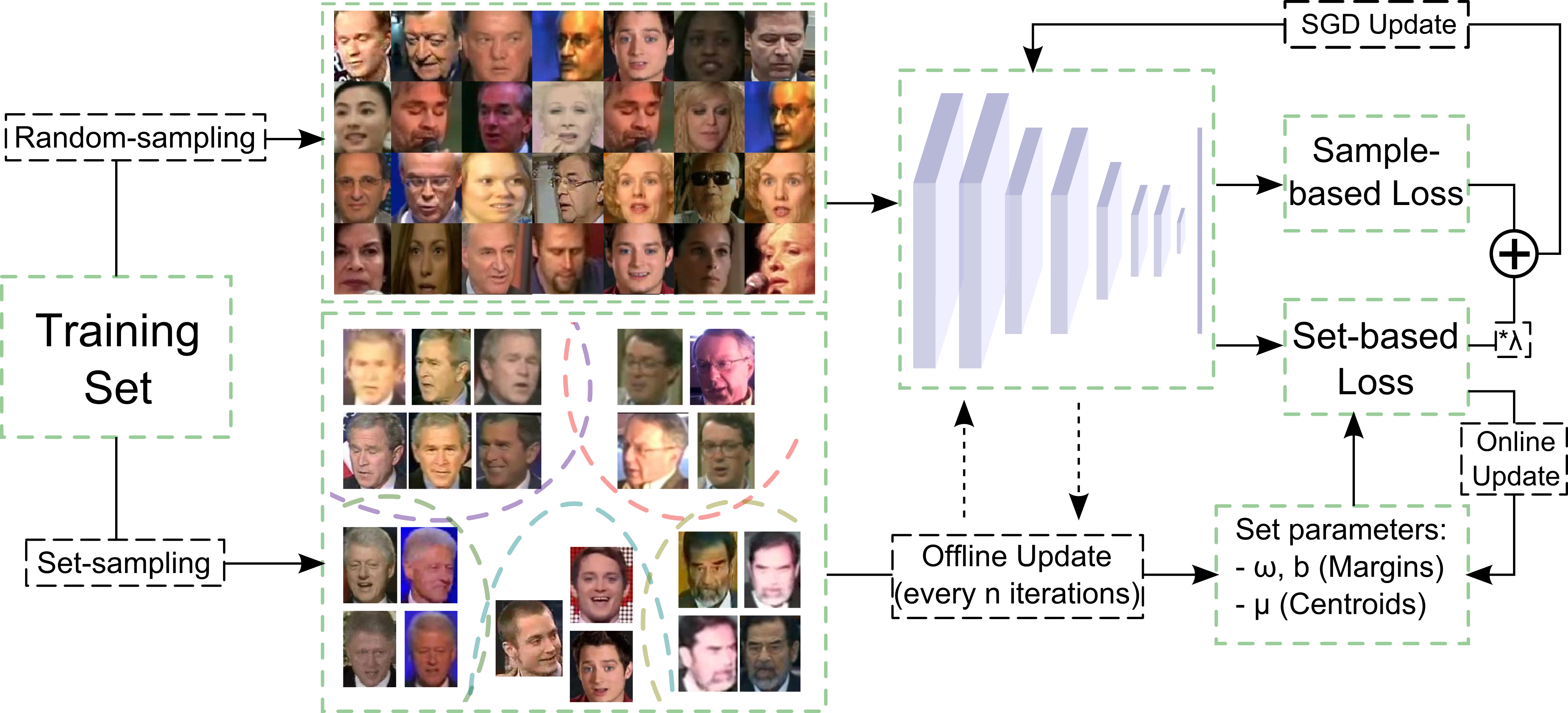}
\end{center}
\caption{Overview of joint sample-based set-based learning. Random face images sampled from the training images in traditional fashion to train a CNN. \textbf{Offline Update:} In every n iterations, set of face images are sampled that consist significant amount of images from each identities and fed into the network while training is paused. Resulting feature vectors are used to calculate set parameters whose way is specific to set-based loss used. \textbf{Online Update:} While the training is going, set parameters are updated with a small weight by the parameters calculated with current random batches.}
\label{fig:overview}
\end{figure*}

%-------------------------------------------------------------------------
\section{Related Work}
While the \textbf{traditional embedding} approaches include Neighbourhood Component Analysis \cite{Goldberger2004}, Large Margin Nearest Neighbour \cite{weinberger2009distance} and Nearest Class Mean \cite{mensink2013distance}, state-of-the-art performances are usually achieved by \textbf{deep DML }networks. Contrastive Loss \cite{hadsell2006dimensionality} is one such approach where the features are learned with supervision of a loss computed with (positive or negative) pairs of samples. Triplet Loss \cite{weinberger2009distance} optimizes the relative difference between a positive and a negative pair. Both functions share the goal to minimize the distances between the samples from the same class and to maximize the distances between the samples from different classes. Several extensions were proposed such as lifted structured embedding \cite{oh2016deep} where an advanced hard sample mining introduced within mini-batches for efficiency, and quadruplet embedding \cite{huang2016local} that employs local similarity awareness.

There have been many methods developed for \textbf{set-based recognition} such as CCA \cite{kim2007discriminative} , Manifold-Manifold Distance \cite{wang2008manifold}, Sparse Approximated Nearest Points \cite{hu2012face}, Simultaneous Feature and Dictionary Learning \cite{lu2014simultaneous}, Discriminant-Analysis on Riemannian Manifold of Gaussian Distribution \cite{wang2015discriminant}. Yet, recent \textbf{set-based deep DML studies} show excellent performance, such as Rippel \etal\cite{rippel2015metric} proposed magnet loss that achieve local discrimination by penalizing class distribution overlap and Feng \etal \cite{feng2016deep} combined set presentations (mean, variance, min, max, vlad features) with hashing in a single network for end-to-end learning of binary code of sets. Wen \etal \cite{wen2016discriminative} did the first attempt to combine sample-based loss functions (\eg softmax, contrastive, triplets) with a set-based term called center loss which minimizes the distance of each sample with its corresponding class center. \note{there is a mix in the previous work. you talk about set-based terms, then traditional CCA and then back to deep learning}

Those studies come with their strategies to compute set parameters (\eg clusters, centroids, margins) on-the-go as well. Rippel \etal\cite{rippel2015metric} pause training periodically to cluster samples on the new feature space, Wen \etal \cite{wen2016discriminative} calculates class centroids with vanilla update with momentum in every iteration. After every iteration of the ongoing learning, feature space is being bended and therefore above approximations should be biased. While the first uses the same cluster indices until next refreshment, in the latter momentum update would lead to aggregation of parameter vectors of different feature spaces. Although they are still good approximations, using both ideas together should yield less biased approximations as we do in our experiments.

Most of above sample- or set-based deep DML studies revolve around learning features by pulling positive samples and pushing negative samples. In fact, more discriminative features can be learned by increasing the inter-class distances without forcing to pull all the samples to the same point (i.e. centroid). Although Wen \etal \cite{wen2016discriminative} claims to learn discriminative features rather than just separable ones, Center loss keep pulling and pushing samples no matter how distinct they are. Although Magnet Loss take care of such intra-class variation tolerance with its multi-cluster models, its sophisticated sampling procedure make it difficult to combine it with sample-based objectives. The proposed \margin, on the other hand, cover these problems by calculating inter-class separating hyperplanes and pushing all the samples to the correct side of the margin accordance with their proximity to the margin. This procedure eventually increase maximum possible margin between sets without distorting the intra-class distribution.

Tang \cite{tang2013deep} attempted to learn with margin-based optimization by minimizing squared hinge loss for classification. Yet, ignoring the weight term($w$) in the differentiation appears to be penalizing only slack variables rather than increasing the maximum achievable margin by SVM. Further, the study is not clear about integration of SVM with SGD in the loss layer. \cite{crosswhite2016template} is another related SVM based study where SVMs are used during the testing time for template adaptation rather than supervising the network to learn better embedding as in our case. 

%-------------------------------------------------------------------------
\section{Proposed Set-based Learning Framework}

In this section, we present our framework to combine sample-based and set-based learning using our novel set-based \margin and two other set-based loss functions similar to the existing works. Let us begin with generalized version of the joint loss formula given by \cite{wen2016discriminative} as following:
\begin{align}
\mathcal{L} = \sum_i \lambda_i \mathcal{L}_{i(Sample-based)} + \sum_j \lambda_j \mathcal{L}_{j(Set-based)}
\end{align}

Sample-based loss functions such as Softmax ($\mathcal{L}_S = - \sum_{i=1}^m log \frac{e^{W^T_{y_i}+b{y_i}}}{\sum^n_{j=1} e^{W^T_{y_i}+b{y_i}}}$) or Triplet are well defined and studied in the literature\cite{schroff2015facenet,parkhi2015deep,sun2014deep}. They often guide a network fed by random batch of input data and without a need of any other information while training.

Beside sample-based supervision, one can benefit from the information extracted by considering set of samples as a whole. Unlike sample-based ones, set-based terms require additional set parameters (\eg linear margin parameters($\omega,b$), centroids($\mu$)) which represent statistics or characteristics of sets. Aggregation of many sample-based and set-based loss terms has a potential of leading to better representation as each may optimize different aspects of the problem.

%Sample based loss functions (\eg Softmax) have shown to be effective to learn separable features in many tasks. Face recognition, however, require more discriminative features to identify or verify unseen individuals without prior knowledge. 

Below we study strategies to extract set statistics and characteristic for set-based learning. Then we introduce a new set-term and review several set-based loss terms similar to the previous studies.

\begin{figure}[t]
\begin{center}

\subfloat[Initial state and the pulling/pushing caused by green hyperplane]{
  \fbox{\includegraphics[width=0.44\linewidth]{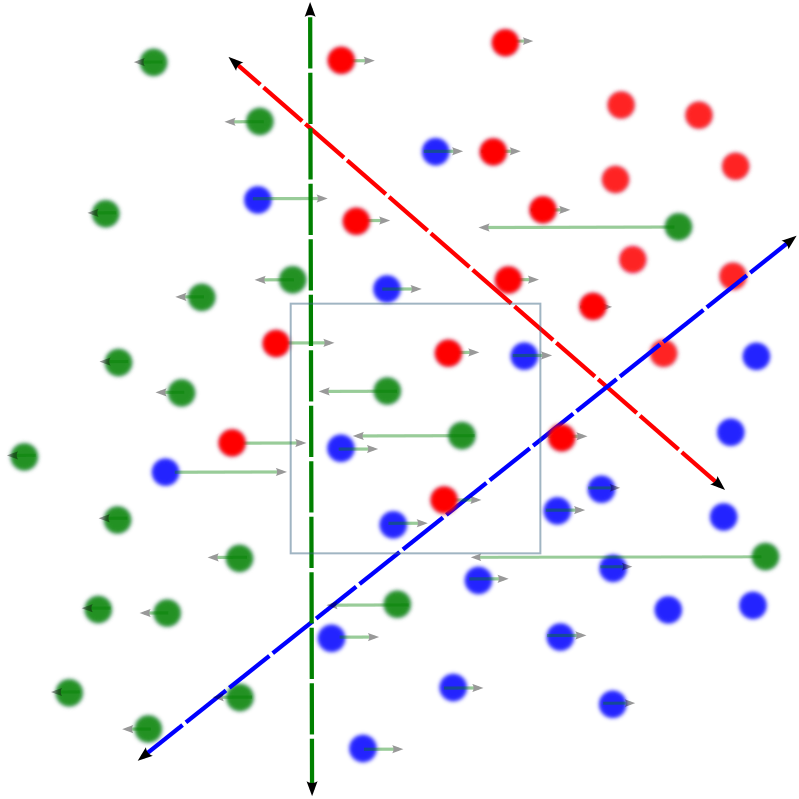}}}
\hspace{0.1in}  
  \subfloat[Zoomed version of (a) with all forces]{
  \fbox{\includegraphics[width=0.44\linewidth]{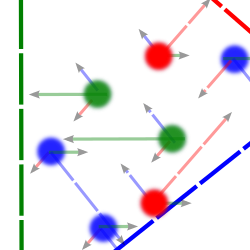}}}\\
  \subfloat[After one update from (a)]{
  \fbox{\includegraphics[width=0.44\linewidth]{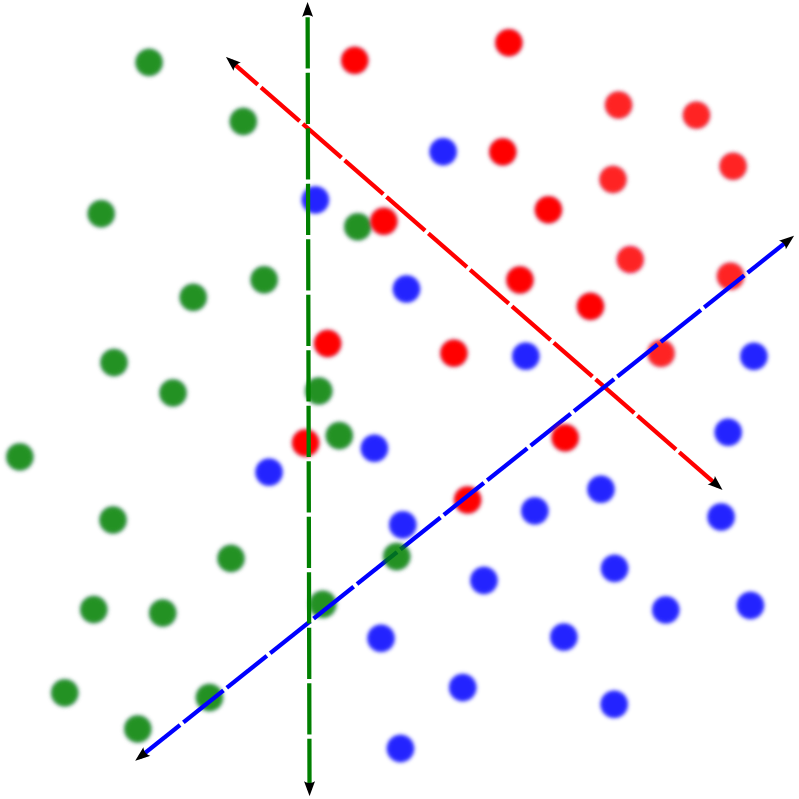}}}
\hspace{0.1in}  
  \subfloat[Convergence]{
  \fbox{\includegraphics[width=0.44\linewidth]{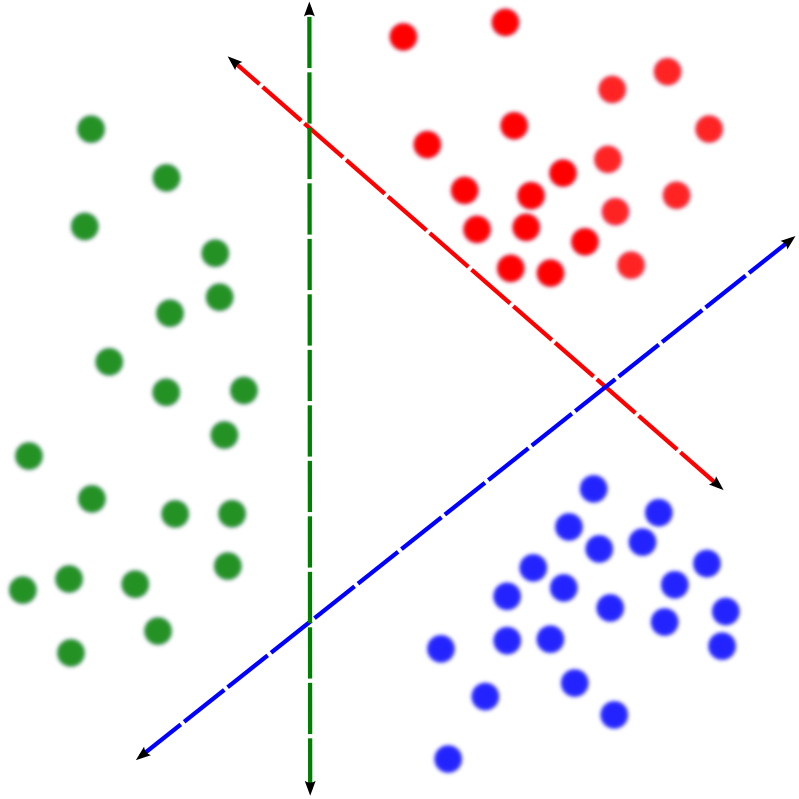}}}
\end{center}
\caption{Initially, \margin requires a good embedding as in (a) to calculate separating hyperplanes. The loss applies to all samples by green plane is indicated with arrows. (c) shows the state after one update for only green plane and (d) at convergence. }
\label{fig:maxmargin}
\end{figure}

\subsection{Set Parameters}
\label{sec:setparams}
Fig. \ref{fig:overview} summarizes joint set- and sample-based learning and shows how set batches and set parameters are operated. As in traditional deep networks, a number random samples (a batch) is fed into the network to compute sample-based loss (\ie Softmax). Then, set-based loss is also computed based on pre-computed set parameters and the weighted sum of their derivatives are backpropagated through the network.

Set parameters are updated periodically in two ways (online and offline) to maintain set-based terms. The best approximation to set parameters can be calculated by sampling a significant number of samples\footnote{We found that 50 images are representative enough} from each identity. Those samples are fed into the network while the network parameters are fixed and set parameters are determined from their features. We call this operation `offline update' which is computationally costly and therefore done in every n iterations.

As training continues, resulting feature space is also changing, thus the set parameters need to be kept on track. `Online update' intends to correct this bias in every iteration by averaging current set parameters with computed set-parameters given the current random batch at the hand. Since number of samples from each class is small, the weight of online-parameters is also small while averaging. While online update keep adapting set parameters to the changing feature space during optimization, offline update periodically correct the biased set parameters caused by mixing parameters of different feature spaces.

\subsection{Set-based Loss Functions}

\subsubsection{\margin}

We propose a novel set-based term, \margin, that maximizes the maximum possible margin between classes. This objective function implicitly pushes all the samples towards correct side of the margin with a vector perpendicular to the hyperplane and a strength exponentially growing towards to negative side of the hyperplane. Even the samples in the correct side of the margin are kept being pushed to increase the maximum margin between the two sets without distorting the intra-class distribution. Fig. \ref{fig:maxmargin} illustrates a synthetic feature space over the iterations supervised by \margin. Given a mini-batch of random $n$ samples uniformly sampled from $m$ classes,  and let the embeddings and the corresponding class labels denoted by $(x_i, y_i)$, the loss and its gradient are computed by the following formulas:

\begin{align}
\mathcal{L}_M = \lambda_M \sum_{i=1}^n \sum_{j=1}^m \dfrac{1}{2\gamma(y_i = j)}e^{- \dfrac{\overline{\delta}(y_i = j)(w_j^Tx_i+b_j)}{||w_j||_2}}\\
\frac{\partial \mathcal{L}_M}{\partial x_i}\! =\!\!\dfrac{\lambda_M}{2n\gamma(y_i = j)}\! \sum_{j=1}^m \! \dfrac{-w_j\overline{\delta}(y_i\!\! =\!\! j)}{||w_j||_2}e^{\!\!- \frac{\overline{\delta}(y_i = j)(w_j^Tx_i+b_j)}{||w_j||_2}}
\end{align}
separating hyperplane for $j$ class is defined as $w_j^Tx+b = 0$. $\overline{\delta}(.)$ and $\gamma(.)$ equal to $1$ if the condition(.) is satisfied and equal to $-1$ and $m-1$ respectively if otherwise. 

Set parameters of \margin ($\omega_j, b_j$) are determined by online and offline updates explained in the previous Section (\ref{sec:setparams}). For offline update, we pause the training in every n($=500$) iterations and after set-sampling, features are extracted from the current state of the network. We then run a support vector machine with linear kernel for each class to obtain best separating hyperplanes in one-against-all manner and save the resulting parameters. Online update is done by running SVM for classes that are represented in the current (random) batch and averaging them with the current parameters with a small weight ($\alpha=0.01$).

\begin{figure}[t]
\begin{center}

  \subfloat[Softmax Loss Alone ($\mathcal{L}_S$ )]{
  \fbox{\includegraphics[width=0.44\linewidth]{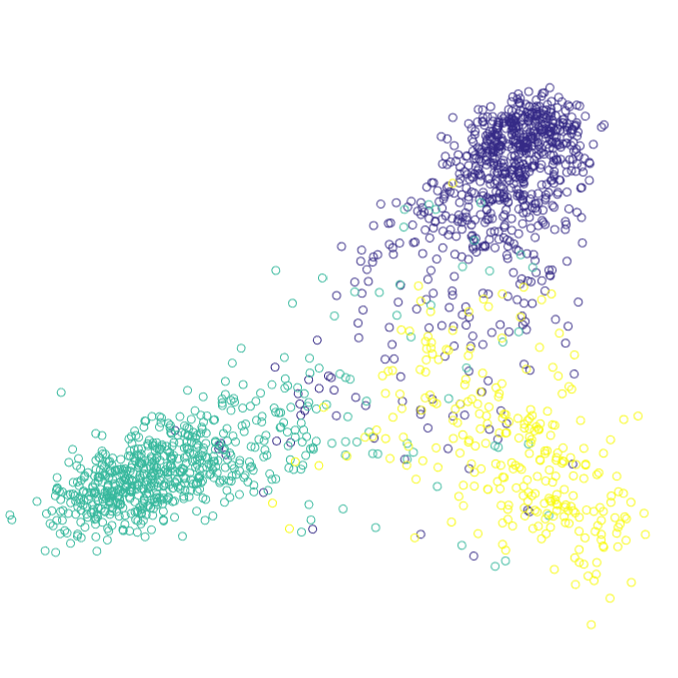}}}\hspace{0.1in}  
    \subfloat[\hspace{-0.02in}Softmax and Center L. ($\!\mathcal{L}_S\!\!+\!\!\mathcal{L}_C\!$)]{
  \fbox{\includegraphics[width=0.44\linewidth]{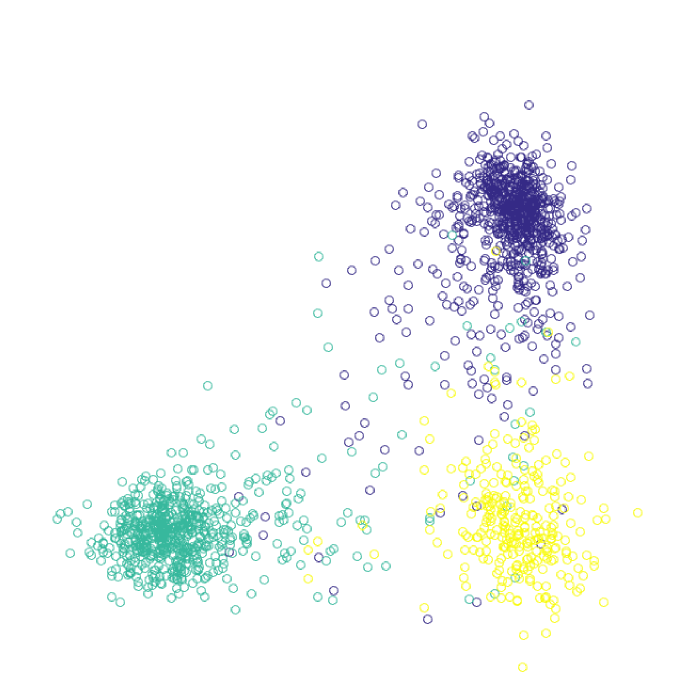}}}\\
    \subfloat[Softmax and Pushing Loss ($\mathcal{L}_S+ \mathcal{L}_P$ )]{
  \fbox{\includegraphics[width=0.44\linewidth]{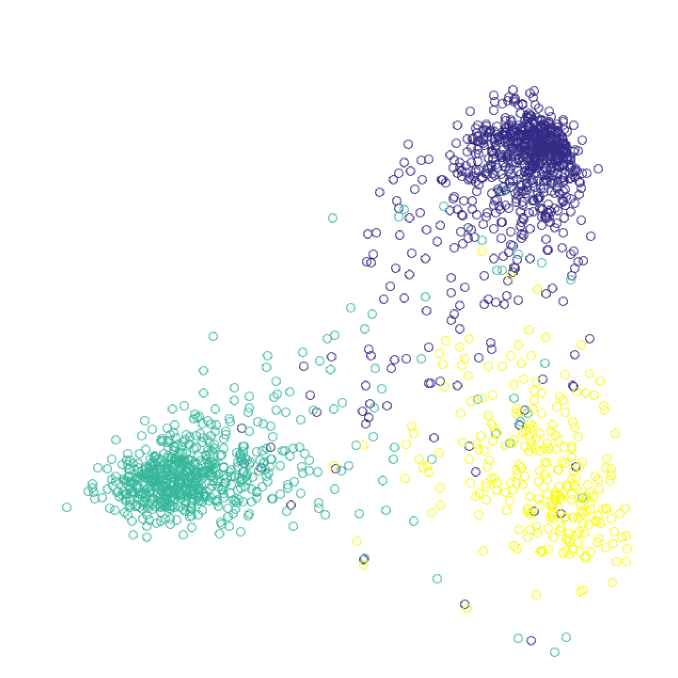}}}\hspace{0.1in}  
    \subfloat[Softmax and \margin ($\mathcal{L}_S+ \mathcal{L}_M$ )]{
  \fbox{\includegraphics[width=0.44\linewidth]{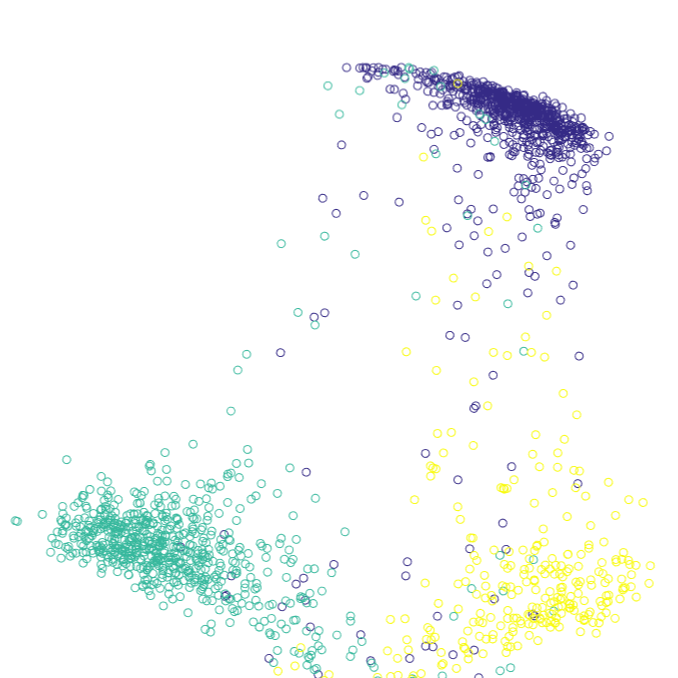}}}
\end{center}
\caption{2D embeddings of images of three learned with different loss functions}
\label{fig:toyexample}
\end{figure}
%-------------------------------------------------------------------------
\subsubsection{Center Loss} Center Loss is proposed by Wen \etal \cite{wen2016discriminative}, minimizes intra-class variety by penalizing distance from samples to their class centroids. This is in fact, a simplified version of the numerator term of Magnet Loss\cite{rippel2015metric} with one cluster. %Unlike the original Center Loss, we refresh center in every epoch to remove the bias caused by momentum update of centroids.

Beside the original definition of the Center loss which includes momentum vanilla update for centroids, we refresh centroids of classes periodically during the training. And unlike the original study \cite{wen2016discriminative}, we introduce Center loss after pretraining the network a while with Softmax alone as having center loss from the beginning would bias deep-net as features and centroids are not yet meaningful.% and that is probably why \cite{wen2016discriminative} had to keep the contribution  parameter ($\lambda = 0.003$) so low.

Center Loss can be defined as below:

\begin{align}
\mathcal{L}_C = \dfrac{\lambda_C}{2} \sum_{i=1}^n ||x_i - c_{y_i} ||_2^2
%\frac{\partial \mathcal{L}_C}{\partial x_i} = \dfrac{\lambda_C}{n}(x_i -  c_{y_i})
\end{align}
where $\lambda_C$ is balancing term and centroids are computed as:

\begin{align}
c_j = \dfrac{\sum_{i=1}^n \delta(y_i = j)x_i}{\sum_{i=1}^n \delta(y_i=j)}
\end{align}
where $\delta(condition)$ equals to $1$ if the condition is satisfied and $0$ otherwise.

%However, instead of clustering samples or updating centroids with momentum over the iterations (as with random batches, each set would have very low number of samples that are not enough for a realistic centroid approximation),

%-------------------------------------------------------------------------
\subsubsection{Pushing Loss} Pushing Loss penalizes very close negative class centroids where penalty decrease exponentially with increasing distance as distant centroids should have much less influence. Here, centroid update and refreshment procedures are kept same as center loss. The formulation of pushing loss is as following:

\vspace{-0.2in}
\begin{align}
\mathcal{L}_P = \dfrac{\lambda_P}{m} \sum_{i=1}^n \sum_{j \neq y_i} e^{-||x_i - c_j ||_2}
\end{align}
%\begin{align}
%\frac{\partial \mathcal{L}_P}{\partial x_i} = \dfrac{\lambda_P}{nm}\dfrac{(c_j - x_i)e^{-||c_j - x_i||_2}}{||c_j - x_i||_2}
%\end{align}

This is also similar to the denominator term of Magnet loss\cite{rippel2015metric}. Although magnet loss contains multiple clusters for attribute concentration, we believe its effect would be minimal for recognition/verification tasks, since deep network should be capable non-linear mapping of multiple clusters into one centroid very easily.

\subsection{Toy Experiments}

We modify our network by setting the dimension of the embedding layer to 2 and train it by supervision of above loss functions using samples of only three people. We plot the embeddings of the samples of individuals with different colors as shown in Figure \ref{fig:toyexample} and observe the contribution of each function. 

Although Fig.\ref{fig:toyexample}(a) shows that softmax alone learn a good representation, it is improved with the contribution of set-based functions. Center Loss and Pushing Loss seems to be functioning similar and \margin looks like providing slightly better separation in terms of identities.

%-------------------------------------------------------------------------
\section{Experiments}

In our experiments, we aim to understand contribution of different set parameter update strategies and compare \margin and other set-based approaches in the same settings.

%-------------------------------------------------------------------------
\subsection{Implementation Details}
\paragraph{Training settings:} We use NNS1 network from \cite{schroff2015facenet} for training which is a reduced version of Google's inception architecture \cite{szegedy2015going}. We increase the dimension of the embedding layer from 128 to 512 and adjust Softmax layer for 2,558 identities. The network is fed with $96 \times 96$ pixel images which are augmented randomly with cropping (between $\%70-\%100$), location, aspect ratio (between $7/8-8/7$), flipping ($0.5$ chance), blurring ($0.5$ chance), brightness, contrast and saturation in every iteration. Input images are linearly scaled to have zero mean and unit norm. SGD is optimized by Adam solver \cite{kingma2014adam} with batch size of 1024 on Nvidia Titan X GPU and we use MatConvNet library \cite{vedaldi2015matconvnet} with a number of modification. We set weight decay to 0.0005 and use batch normalization to avoid overfitting. Training is started with a learning rate of 0.001 and divided by 10 at the 15th and 25th epochs and stopped at 30th epoch.

\paragraph{Training Data:} We use non-aligned and curated version of VGG Face dataset \cite{parkhi2015deep} which consist of around 1M face images of 2,558 individuals who are not included in Youtube Faces (YTF) \cite{wolf2011face} and Labeled Faces in the Wild (LFW) \cite{huang2007labeled} benchmark datasets. Unfortunately, the dataset is publicised by means web links, thus some samples are missing due to broken links. We end up training with reduced version of VGG dataset that consists 0.83M training samples from 2,558 identities.

\paragraph{Testing:} We evaluated performance of \margin and other functions on commonly used YTF and LFW datasets. For both, we follow the defined protocol for the restricted settings with external training. After training, we kept the models fixed and tested on those datasets without further training unlike \cite{parkhi2015deep}. Images are aligned as provided in YTF and for LFW, deep funneling \cite{Huang2012a} is used for alignment. We use embedding layer output of each image as representation and average features of frames from the same videos (only for YTF). Similarity between pairs of images or videos is computed by cosine distance of mean feature vectors.

%-------------------------------------------------------------------------
\subsection{Balancing Term Tuning}
\begin{figure}
\begin{center}

\vspace{-0.2in}
\includegraphics[width=0.9\linewidth]{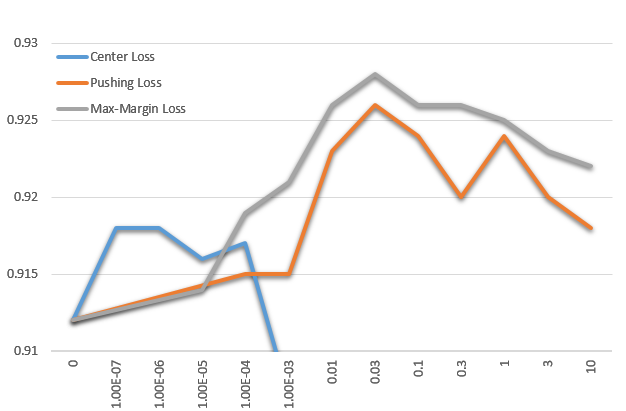}

\caption{Fine-tuning balancing term $\lambda$}
\label{fig:lambda}
\end{center}
\end{figure}

\begin{figure*}[t]
\begin{center}

  \subfloat[Accuracies on YTF]{
  \fbox{\includegraphics[width=0.3\linewidth]{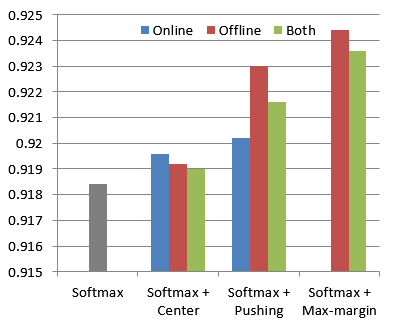}}}
  \subfloat[AUC on YTF]{
  \fbox{\includegraphics[width=0.3\linewidth]{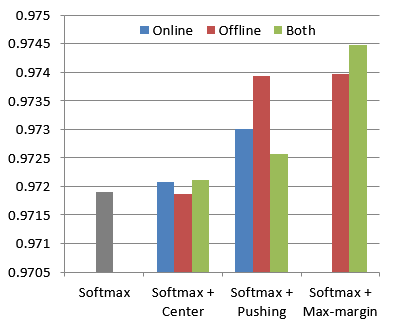}}}
  \subfloat[100$\%$- EER on YTF]{
  \fbox{\includegraphics[width=0.3\linewidth]{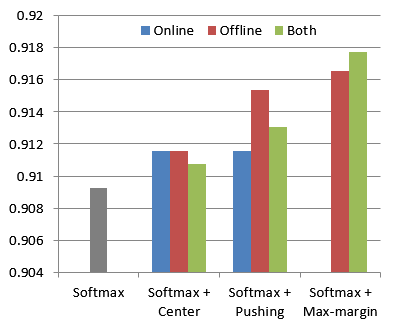}}}
  
  \hspace{0.02in}    
  \subfloat[Accuracies on LFW]{
  \fbox{\includegraphics[width=0.3\linewidth]{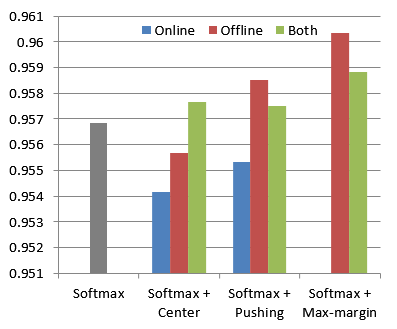}}}
  \subfloat[AUC on LFW]{
  \fbox{\includegraphics[width=0.3\linewidth]{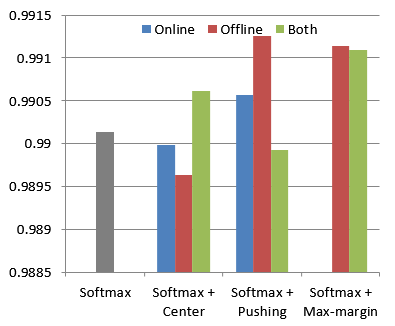}}}
  \subfloat[100$\%$- EER on LFW]{
  \fbox{\includegraphics[width=0.3\linewidth]{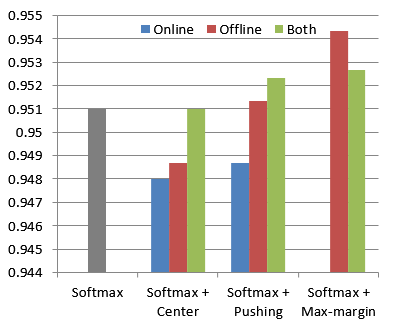}}}
  
\end{center}
\caption{Comparison of the three set-based loss function with online, offline update strategies for set parameters.}

\label{fig:update}
\end{figure*}
Despite motivations using set-based terms, sample-based terms are necessary to stabilize supervision as set characteristics may not be always fully presented in the feature space. Therefore, we train our networks to obtain good features for the first 15 epochs with only supervision of Softmax.

After keeping the pretrained model fixed, we combine set-based loss functions with a balancing term ($\lambda$) which is fine-tuned on a small subset\footnote{One fifth of YTF data set}. According to the Figure \ref{fig:lambda}, we fix $\lambda$ parameters to  $\lambda_M = 0.03, \lambda_P = 0.03, \lambda_C = 0.0001$ in the rest of the experiments.

\subsection{Effectiveness of Online-Offline Updates}
\begin{table}
\begin{center}
\begin{tabular}{|c|l|c|c|c|c|c|c|}
\hline
 $\lambda_C$ & Update & AUC & Acc. & 100$\%$- EER \\

\hline

 0.003& Online & 98.98 &95.45 &94.80 \\
%YTF 0.0276239848330489	0.0808000000000000	0.0891773758115053
%LFW 0.0101888888888884	0.0455000000000000	0.0520000000000000
 0.0001& Online & 99.00 &95.41 &94.80 \\

%YTF 0.0279234137127544	0.0804000000000000	0.0884179052479415
%LFW 0.0100166666666664	0.0458333333333333	0.0520000000000000
 0.003& Both & 98.91
 &95.43
 &94.80 \\
 
0.0001& Both & 99.06
 &95.77
 &95.10 \\

\hline
\end{tabular}
\end{center}
\caption{Performance of Center Loss ($\mathcal{L}_S + \mathcal{L}_C$) on LFW dataset under different settings. The settings used in the original Center Loss paper \cite{wen2016discriminative} (first line) is gradually improved in our experiments.}
\label{tab:centerloss}
\end{table}

In order to justify the small changes we made with Center Loss, we have done some controlled experiments to observe effect of using offline update and new finetuned $\lambda_C$ parameter. As can be seen in Table \ref{tab:centerloss} Each seems to be contributing slightly ending up with around $\%0.3$ improvement on LFW dataset.

Figure \ref{fig:update} show comparison of \margin with other loss functions with different update rules. Although we see close performances, \margin show slightly better performance over other set-based terms. Effect of different update strategies show that offline update alone show consistently better performances which is not expected. The reason could be online set parameter update is done with too big $\alpha$ parameter. We conclude that further investigation of $\alpha$ parameter is needed

%-------------------------------------------------------------------------
\subsection{Benchmark Performances}

\begin{table*}
\begin{center}
\begin{tabular}{|l|c|c|c|c|c|c|c|}
\hline
Method & \begin{tabular}{@{}c@{}}$\#$Training\\ Images\end{tabular} & $\#$Ids & \begin{tabular}{@{}c@{}}Input\\ Size\end{tabular} & \begin{tabular}{@{}c@{}}Network\\($\#$Params.)\end{tabular} & \begin{tabular}{@{}c@{}}FT on\\YTF or LFW\end{tabular}& \begin{tabular}{@{}c@{}}Accuracy\\on YTF ($\%$)\end{tabular} & \begin{tabular}{@{}c@{}}Accuracy\\on LFW ($\%$)\end{tabular}  \\
\hline\hline

DeepFace \cite{taigman2014deepface} & 4.4M & 4,030 & 152$\times$152& AlexNet(120M) & No & 91.4 & 97.35  \\

VGG Face \cite{parkhi2015deep} & 2.62M & 2,622 & 224$\times$224& VGG(138M) &Yes & 97.3 & 98.95 \\

VGG Face \cite{parkhi2015deep} & 2.62M & 2,622 & 224$\times$224& VGG(138M)& No & 91.6 & - \\

\hline
\boldmath{$\mathcal{L}_S + \mathcal{L}_M$}& 0.83M & 2,558 & 96$\times$96 & NNS1(26M) & No &\textbf{92.44} &\textbf{96.03} \\
\hline
\end{tabular}
\end{center}
\caption{Verification performance comparison of different loss functions and methods on YTF dataset. Note that our method uses fairly less training samples with lower input size on a shallower network. Further we do not fine-tune our network on the target datasets like \cite{parkhi2015deep}}
\label{tab:ytf_ver}
\end{table*}

%\begin{table*}
%\begin{center}
%\begin{tabular}{|l|c|c|c|c|c|c|c|c|}
%\hline
%Method & \begin{tabular}{@{}c@{}}$\#$Training\\ Images\end{tabular} & $\#$Ids & \begin{tabular}{@{}c@{}}Input\\ Size\end{tabular} & \begin{tabular}{@{}c@{}}Network\\($\#$Params.)\end{tabular} & \begin{tabular}{@{}c@{}}FT on\\LFW\end{tabular}& AUC & Accuracy & \begin{tabular}{@{}c@{}}100$\%$\\- EER\end{tabular}  \\
%\hline\hline
%
%DeepFace \cite{taigman2014deepface} & 4.4M & 4,030 & 152$\times$152& AlexNet(120M) & No & - &97.35 &-  \\
%
%FaceNet \cite{schroff2015facenet} & 200M & 8M & 220$\times$220& NN1(140M) & - & - &98,87 &-  \\
%
%VGG Face \cite{parkhi2015deep} & 2.62M & 2,622 & 224$\times$224& VGG(138M) &Yes & - &98,95 &-  \\
%
%Center L. \cite{wen2016discriminative} & 0.7M & 17,189& 112$\times$96& LC($\>$1M) & No & - &99,28 &- \\
%
%\hline
%$\mathcal{L}_S$ & 0.83M & 2,558 & 96$\times$96 & NNS1(26M) & No &99,01
% &95,68
% &95,10
% \\
%\boldmath{$\mathcal{L}_S + \mathcal{L}_M$}& 0.83M & 2,558 & 96$\times$96 & NNS1(26M) & No &\textbf{99,11}
% &\textbf{96,03}
% &\textbf{95,43}
% \\
%$\mathcal{L}_S + \mathcal{L}_P$& 0.83M & 2,558 & 96$\times$96 & NNS1(26M) & No & 99,12
% &95,87
%&95,43
% \\
%$\mathcal{L}_S + \mathcal{L}_C$& 0.83M & 2,558 & 96$\times$96 & NNS1(26M) & No & 99,06
% &95,77
% &95,10
% \\
%$\mathcal{L}_S$ + $\mathcal{L}_P$ + $\mathcal{L}_M$& 0.83M & 2,558 & 96$\times$96 & NNS1(26M) & No & 99,12
% &95,90
% &95,30
% \\
%\hline
%\end{tabular}
%\end{center}
%\caption{Verification performance comparison of different loss functions and methods on LFW dataset}
%\label{tab:lfw_ver}
%\end{table*}

Verification performance results of the proposed \margin is compared with the other state-of-the-art methods in Table \ref{tab:ytf_ver} and other set-based functions in Figure \ref{fig:update} for LFW and YTF datasets. Among set-based terms, we obtain the best performance with \margin where we improve $\%0.35-0.6$ over Softmax. The improvement seems to be small as we add set terms after pre-training with only softmax for 15 epochs which we have the 10 and 100 times lower learning rate. Yet, we are not interested achieving state-of-the-art performance as long as we can compare different set-terms in a meaningful experimental settings.

While comparing with other methods, one should also notice the differences in number of training images, identities, input size and network specifications. Most of those studies have such advantages and not directly comparable with our results. For example, VGG Face\cite{parkhi2015deep} yields a significant improvement in the results when they further train (FT) their network on the test set (YTF and LFW) with cross validation. Our baseline Softmax Loss achieve similar accuracy with VGG Face\cite{parkhi2015deep} without FT, although VGG Face is trained with around 3 times larger training set, 5 times bigger input size and 5 times deeper network. While our baseline score is around such state-of-the-art study, we can argue that including such set-based terms would improve their results as well as our framework is compatible with their designs.

\section{Conclusion}

This paper studies joint sample-based and set-based embedding learning for face recognition. We review different set terms in the literature and propose nove \margin. We also explain strategies to maintain set-based learning during training.

Our results show the contribution of different terms and validity of the proposed set-based function which yields slight improvement over softmax baseline. Further experiments give us better insight about set-based learning methods. Without aiming is still an ongoing work.

\section*{Acknowledgements}
This work was supported in part by the EPSRC Programme
Grant `FACER2VM' (EP/N007743/1) and Baris Gecer is funded by the Turkish Ministry of National Education.

%\begin{figure}
%\begin{tabular}{ccc}
%\bmvaHangBox{\fbox{\parbox{2.7cm}{~\\[2.8mm]
%\rule{0pt}{1ex}\hspace{2.24mm}\includegraphics[width=2.33cm]{images/eg1_largeprint.png}\\[-0.1pt]}}}&
%\bmvaHangBox{\fbox{\includegraphics[width=2.8cm]{images/eg1_largeprint.png}}}&
%\bmvaHangBox{\fbox{\includegraphics[width=5.6cm]{images/eg1_2up.png}}}\\
%(a)&(b)&(c)
%\end{tabular}
%\caption{}
%\label{fig:teaser}
%\end{figure}

%\begin{figure*}
%\begin{center}
%\fbox{\rule{0pt}{2in} \rule{.9\linewidth}{0pt}}
%\end{center}
%   \caption{Example of a short caption, which should be centered.}
%\label{fig:short}
%\end{figure*}

%\begin{table}
%\begin{center}
%\begin{tabular}{|l|c|}
%\hline
%Method & Frobnability \\
%\hline\hline
%Theirs & Frumpy \\
%Yours & Frobbly \\
%Ours & Makes one's heart Frob\\
%\hline
%\end{tabular}
%\end{center}
%\caption{Results.   Ours is better.}
%\end{table}

{\small
\bibliographystyle{ieee}
\bibliography{egbib}

\begin{thebibliography}{10}\itemsep=-1pt

\bibitem{crosswhite2016template}
N.~Crosswhite, J.~Byrne, O.~M. Parkhi, C.~Stauffer, Q.~Cao, and A.~Zisserman.
\newblock Template adaptation for face verification and identification.
\newblock {\em arXiv preprint arXiv:1603.03958}, 2016.

\bibitem{donahue2014decaf}
J.~Donahue, Y.~Jia, O.~Vinyals, J.~Hoffman, N.~Zhang, E.~Tzeng, and T.~Darrell.
\newblock Decaf: A deep convolutional activation feature for generic visual
  recognition.
\newblock In {\em Icml}, volume~32, pages 647--655, 2014.

\bibitem{feng2016deep}
J.~Feng, S.~Karaman, I.~Jhuo, S.-F. Chang, et~al.
\newblock Deep image set hashing.
\newblock {\em arXiv preprint arXiv:1606.05381}, 2016.

\bibitem{gecer2016detection}
B.~Gecer.
\newblock Detection and classification of breast cancer in whole slide
  histopathology images using deep convolutional networks.
\newblock {\em Diss. Bilkent University}, 2016.

\bibitem{Goldberger2004}
J.~Goldberger, S.~Roweis, G.~Hinton, and R.~Salakhutdinov.
\newblock Neighbourhood components analysis.
\newblock In {\em Proceedings of the 17th International Conference on Neural
  Information Processing Systems}, NIPS'04, pages 513--520, Cambridge, MA, USA,
  2004. MIT Press.

\bibitem{hadsell2006dimensionality}
R.~Hadsell, S.~Chopra, and Y.~LeCun.
\newblock Dimensionality reduction by learning an invariant mapping.
\newblock In {\em Computer vision and pattern recognition, 2006 IEEE computer
  society conference on}, volume~2, pages 1735--1742. IEEE, 2006.

\bibitem{hu2012face}
Y.~Hu, A.~S. Mian, and R.~Owens.
\newblock Face recognition using sparse approximated nearest points between
  image sets.
\newblock {\em IEEE Transactions on Pattern Analysis and Machine Intelligence},
  34(10):1992--2004, 2012.

\bibitem{huang2016local}
C.~Huang, C.~C. Loy, and X.~Tang.
\newblock Local similarity-aware deep feature embedding.
\newblock In {\em Advances in Neural Information Processing Systems}, pages
  1262--1270, 2016.

\bibitem{Huang2012a}
G.~B. Huang, M.~Mattar, H.~Lee, and E.~Learned-Miller.
\newblock Learning to align from scratch.
\newblock In {\em NIPS}, 2012.

\bibitem{huang2007labeled}
G.~B. Huang, M.~Ramesh, T.~Berg, and E.~Learned-Miller.
\newblock Labeled faces in the wild: A database for studying face recognition
  in unconstrained environments.
\newblock Technical report, Technical Report 07-49, University of
  Massachusetts, Amherst, 2007.

\bibitem{kim2007discriminative}
T.-K. Kim, J.~Kittler, and R.~Cipolla.
\newblock Discriminative learning and recognition of image set classes using
  canonical correlations.
\newblock {\em IEEE Transactions on Pattern Analysis and Machine Intelligence},
  29(6), 2007.

\bibitem{kingma2014adam}
D.~Kingma and J.~Ba.
\newblock Adam: A method for stochastic optimization.
\newblock {\em arXiv preprint arXiv:1412.6980}, 2014.

\bibitem{lecun2015deep}
Y.~LeCun, Y.~Bengio, and G.~Hinton.
\newblock Deep learning.
\newblock {\em Nature}, 521(7553):436--444, 2015.

\bibitem{lu2014simultaneous}
J.~Lu, G.~Wang, W.~Deng, and P.~Moulin.
\newblock Simultaneous feature and dictionary learning for image set based face
  recognition.
\newblock In {\em European Conference on Computer Vision}, pages 265--280.
  Springer, 2014.

\bibitem{mensink2013distance}
T.~Mensink, J.~Verbeek, F.~Perronnin, and G.~Csurka.
\newblock Distance-based image classification: Generalizing to new classes at
  near-zero cost.
\newblock {\em IEEE transactions on pattern analysis and machine intelligence},
  35(11):2624--2637, 2013.

\bibitem{oh2016deep}
H.~Oh~Song, Y.~Xiang, S.~Jegelka, and S.~Savarese.
\newblock Deep metric learning via lifted structured feature embedding.
\newblock In {\em Proceedings of the IEEE Conference on Computer Vision and
  Pattern Recognition}, pages 4004--4012, 2016.

\bibitem{parkhi2015deep}
O.~M. Parkhi, A.~Vedaldi, and A.~Zisserman.
\newblock Deep face recognition.
\newblock In {\em BMVC}, volume~1, page~6, 2015.

\bibitem{qian2015fine}
Q.~Qian, R.~Jin, S.~Zhu, and Y.~Lin.
\newblock Fine-grained visual categorization via multi-stage metric learning.
\newblock In {\em Proceedings of the IEEE Conference on Computer Vision and
  Pattern Recognition}, pages 3716--3724, 2015.

\bibitem{rippel2015metric}
O.~Rippel, M.~Paluri, P.~Dollar, and L.~Bourdev.
\newblock Metric learning with adaptive density discrimination.
\newblock {\em arXiv preprint arXiv:1511.05939}, 2015.

\bibitem{schroff2015facenet}
F.~Schroff, D.~Kalenichenko, and J.~Philbin.
\newblock Facenet: A unified embedding for face recognition and clustering.
\newblock In {\em Proceedings of the IEEE Conference on Computer Vision and
  Pattern Recognition}, pages 815--823, 2015.

\bibitem{sharif2014cnn}
A.~Sharif~Razavian, H.~Azizpour, J.~Sullivan, and S.~Carlsson.
\newblock Cnn features off-the-shelf: an astounding baseline for recognition.
\newblock In {\em Proceedings of the IEEE Conference on Computer Vision and
  Pattern Recognition Workshops}, pages 806--813, 2014.

\bibitem{snoek2015scalable}
J.~Snoek, O.~Rippel, K.~Swersky, R.~Kiros, N.~Satish, N.~Sundaram, M.~M.~A.
  Patwary, M.~Prabhat, and R.~P. Adams.
\newblock Scalable bayesian optimization using deep neural networks.
\newblock In {\em ICML}, pages 2171--2180, 2015.

\bibitem{sun2014deep}
Y.~Sun, X.~Wang, and X.~Tang.
\newblock Deep learning face representation from predicting 10,000 classes.
\newblock In {\em Proceedings of the IEEE Conference on Computer Vision and
  Pattern Recognition}, pages 1891--1898, 2014.

\bibitem{szegedy2015going}
C.~Szegedy, W.~Liu, Y.~Jia, P.~Sermanet, S.~Reed, D.~Anguelov, D.~Erhan,
  V.~Vanhoucke, and A.~Rabinovich.
\newblock Going deeper with convolutions.
\newblock In {\em Proceedings of the IEEE conference on computer vision and
  pattern recognition}, pages 1--9, 2015.

\bibitem{taigman2014deepface}
Y.~Taigman, M.~Yang, M.~Ranzato, and L.~Wolf.
\newblock Deepface: Closing the gap to human-level performance in face
  verification.
\newblock In {\em Proceedings of the IEEE conference on computer vision and
  pattern recognition}, pages 1701--1708, 2014.

\bibitem{tang2013deep}
Y.~Tang.
\newblock Deep learning using linear support vector machines.
\newblock {\em arXiv preprint arXiv:1306.0239}, 2013.

\bibitem{vedaldi2015matconvnet}
A.~Vedaldi and K.~Lenc.
\newblock Matconvnet: Convolutional neural networks for matlab.
\newblock In {\em Proceedings of the 23rd ACM international conference on
  Multimedia}, pages 689--692. ACM, 2015.

\bibitem{wang2008manifold}
R.~Wang, S.~Shan, X.~Chen, and W.~Gao.
\newblock Manifold-manifold distance with application to face recognition based
  on image set.
\newblock In {\em Computer Vision and Pattern Recognition, 2008. CVPR 2008.
  IEEE Conference on}, pages 1--8. IEEE, 2008.

\bibitem{wang2015discriminant}
W.~Wang, R.~Wang, Z.~Huang, S.~Shan, and X.~Chen.
\newblock Discriminant analysis on riemannian manifold of gaussian
  distributions for face recognition with image sets.
\newblock In {\em Proceedings of the IEEE Conference on Computer Vision and
  Pattern Recognition}, pages 2048--2057, 2015.

\bibitem{weinberger2009distance}
K.~Q. Weinberger and L.~K. Saul.
\newblock Distance metric learning for large margin nearest neighbor
  classification.
\newblock {\em Journal of Machine Learning Research}, 10(Feb):207--244, 2009.

\bibitem{wen2016discriminative}
Y.~Wen, K.~Zhang, Z.~Li, and Y.~Qiao.
\newblock A discriminative feature learning approach for deep face recognition.
\newblock In {\em European Conference on Computer Vision}, pages 499--515.
  Springer, 2016.

\bibitem{wolf2011face}
L.~Wolf, T.~Hassner, and I.~Maoz.
\newblock Face recognition in unconstrained videos with matched background
  similarity.
\newblock In {\em Computer Vision and Pattern Recognition (CVPR), 2011 IEEE
  Conference on}, pages 529--534. IEEE, 2011.

\end{thebibliography}
}

\end{document}